\documentclass{article}

\usepackage[main, final]{neurips_2025}

\usepackage[utf8]{inputenc} 
\usepackage[T1]{fontenc}    
\usepackage{hyperref}       
\usepackage{url}            
\usepackage{booktabs}       
\usepackage{amsfonts}       
\usepackage{nicefrac}       
\usepackage{microtype}      
\usepackage{xcolor}         
\usepackage[table]{xcolor}
\usepackage{url}
\usepackage[utf8]{inputenc}

\usepackage{amsmath}
\usepackage{amssymb}
\usepackage{multirow}
\usepackage{booktabs}

\usepackage[utf8]{inputenc}
\usepackage[most]{tcolorbox} 
\usepackage{cleveref}

\usepackage{fontawesome5} 

\usepackage{fancyhdr}
\usepackage{pifont}
\usepackage{fontawesome5}

\usepackage{fixltx2e}  
\usepackage{xeCJK}

\usepackage{newtxtext,newtxmath} 
\usepackage[most]{tcolorbox}
\usepackage{xcolor}
\usepackage{subcaption}

\usepackage{graphicx}  
\usepackage{tikz}
\usetikzlibrary{arrows.meta, positioning, calc, fit, backgrounds, shapes.geometric}

\makeatletter
\renewcommand{\@neuripsordinal}{}
\newcommand{\@neuripsyear}{2025}
\makeatother

\definecolor{BoxBg}{RGB}{253, 253, 253}
\definecolor{TitleBg}{RGB}{245, 245, 245}
\definecolor{FrameColor}{RGB}{120, 120, 120}

\newtcolorbox{promptbox}[1]{
    enhanced,
    colback=BoxBg,
    colframe=FrameColor,
    boxrule=0.5pt,
    sharp corners,
    top=4mm, bottom=3mm, left=4mm, right=4mm,
    fonttitle=\normalfont,        
    coltitle=black,
    colbacktitle=TitleBg,
    title={#1},
    attach boxed title to top left={yshift=-2.5mm, xshift=3mm},
    boxed title style={
        boxrule=0.5pt,
        colframe=FrameColor,
        sharp corners,
    }
}

\newcommand{\token}[1]{{\texttt{\small <|#1|>}}}

\title{
Pailitao-MMSearch: Building Native E-Commerce Multimodal Search Foundation
  } 

\author{
    \begin{tabular}{c}
        Xiaohan Ye,\; Xu Chen\footnotemark[1], \; Zihan Gong, \; Jian Ding, \; Lianyu Du, \\
        Baicheng Chen,  Yunmeng Shu, 
        Jingqian Zhao, Zhixiang Zhao, Shuaiqi Jia, Chong Ma, \\ Shuwen Xiao, Xiangheng Kong, Yuan Gao, Jun Song, Jinsong Lan, Xiaoyong Zhu, Bo Zheng \\
        \quad \\
        Taobao \& Tmall Group of Alibaba \\
        {\small \texttt{
        \{yxh268746,huaisong.cx\}@alibaba-inc.com
        }} \\
        {\small \texttt{
        yexiaohan@whu.edu.cn,xuchen2016@sjtu.edu.cn
        }}
    \end{tabular}
}

\begin{document}

\begin{tikzpicture}[remember picture, overlay]                                               
      \node[anchor=north west, inner sep=0pt] at ([xshift=3.8cm, yshift=-1.8cm]current page.north  
  west)                                                                                        
          {\includegraphics[height=1.2cm]{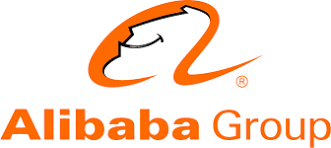}};                          
  \end{tikzpicture} 

\maketitle
\renewcommand{\thefootnote}{\fnsymbol{footnote}}

\renewcommand{\thefootnote}{\fnsymbol{footnote}}
\footnotetext[1]{Corresponding Author and Team Leader of Pailitao's Multi-Modal Ranking and Relevance.}

\begin{abstract}
The evolution of e-commerce has fundamentally transformed how users search for products, shifting from simple text-based keyword queries to complex multimodal interactions that seamlessly combine product images, natural language descriptions, and mixed-intent instructions. However, existing approaches face a critical dilemma: single-modal specialist models, deployed independently for text retrieval, visual search, and voice recognition, operate in isolation and cannot handle cross-modal queries, while general-purpose vision-language models lack the domain-specific knowledge necessary for fine-grained product understanding, user behavior modeling, and commercial intent reasoning.
In this work, we present \textbf{Pailitao-MMSearch}, one native e-commerce multimodal search foundation model designed to bridge this gap. Our approach introduces three key innovations:
(1)~\textbf{HybSID} (Hybrid Semantic ID), a novel product tokenization scheme that unifies discrete semantic codes with continuous multimodal embeddings, enabling both coarse-grained product generalization and fine-grained discrimination within a single autoregressive language model;
(2)~a \textbf{two-stage continual pre-training} strategy that first injects e-commerce domain knowledge, encompassing product attribute understanding, cross-modal alignment and user behavior patterns, and subsequently preserves the base model's natural language reasoning capabilities through on-policy distillation (OPD), addressing the notorious catastrophic forgetting problem without access to the original pre-training corpus;
and (3)~a \textbf{hybrid reasoning post-training} pipeline that integrates multi-task supervised fine-tuning across diverse e-commerce search scenarios, difficulty-aware chain-of-thought reasoning for complex intents, reinforcement learning with verifiable product-grounded rewards, and multi-expert knowledge fusion.
Built upon Qwen and deployed on Taobao's Pailitao multimodal search platform, Pailitao-MMSearch achieves substantial improvements in online A/B testing, including up to \textbf{+13.61\%} in Gross Merchandise Volume (GMV) and \textbf{+8.21\%} in transaction volume compared to traditional multi-modal search pipeline, demonstrating the effectiveness of our native e-commerce multimodal search large language models.

\end{abstract}

\newpage

\vspace{0.3cm}
\tableofcontents 
\newpage         

\section{Introduction} \label{intro}
The landscape of e-commerce product search has undergone a fundamental transformation in recent years. As mobile platforms mature and artificial intelligence technologies advance, users increasingly rely on multimodal interactions to express their shopping intent. They usually photograph a product encounter in daily life and annotate it with textual modifications, combining reference images with free-form text descriptions, or issue compound queries that simultaneously specify visual attributes and functional requirements. This behavioral shift from single-modal keyword search to rich, multimodal query expressions places unprecedented demands on the underlying search infrastructure. 


Despite the clear demand, current production systems remain ill-equipped to handle such multimodal search scenarios. The prevailing industrial architectures fall into two categories, each with critical structural limitations:

\textbf{Single-modal specialist models.} Single-modal specialist retrieval where queries and candidates are matched within a single modality, has been the dominant paradigm in e-commerce search for decades. On the textual side, dual-encoder architectures and dense retrieval methods~\citep{dssm,dpr,simcse,e5} map queries and product titles into a shared embedding space for efficient nearest-neighbor matching. On the visual side, deep metric learning and product embedding techniques~\citep{ebay_visual_search,pinterest_visual_search,itemsage,eproduct} enable image-to-image retrieval for scenarios such as ``find similar products.'' In practice, most large-scale e-commerce platforms deploy these text-based and image-based models independently~\citep{tdm,din}, alongside separate voice-based query understanding modules. These models are trained independently on modality-specific objectives, producing isolated representation spaces that cannot be composed at inference time. When a user issues a multimodal query combining image and text, the system must resort to ad-hoc fusion heuristics---typically decomposing the query into single-modal sub-queries, routing each to the corresponding specialist model, and merging results through handcrafted rules or post-processing pipelines. This ``patchwork'' approach introduces substantial system complexity, increases end-to-end latency, and fundamentally cannot capture the joint semantics of cross-modal queries where the meaning emerges from the \emph{interaction} between modalities rather than their independent interpretations.

\textbf{General-purpose vision-language models (VLMs).} The rapid progress in multimodal foundation models~\citep{liu2024llava,qwen2vl,qwen3,gpt4} offers a seemingly natural solution: leverage a single model capable of processing both images and text. However, these general-purpose VLMs are pre-trained on broad web-scale data and lack critical e-commerce domain knowledge. Specifically, they exhibit three key deficiencies in commercial search scenarios: (i)~\emph{lack of e-commerce domain understanding}---general-purpose VLMs have no knowledge of domain-specific product taxonomies, attribute ontologies, or commercial semantics. They cannot accurately interpret fine-grained product properties such as material composition, functional specifications, or brand-tier distinctions that are critical for personalized purchase decisions;
(ii)~\emph{absence of collaborative filtering signals}---they have no awareness of user preference patterns, purchase co-occurrence statistics, or personalized ranking signals that are essential for relevance in e-commerce; and (iii)~\emph{inability to generate structured product references}---they cannot directly produce product identifiers or retrieve from a catalog of billions of items, requiring additional retrieval modules that break the end-to-end paradigm.

These limitations manifest as concrete user experience failures in production. Our analysis of online search logs reveals systematic patterns: models ignoring critical textual instructions while attending only to image content (e.g., returning similar dresses when the user explicitly requests ``jumpsuit''), failing to leverage image context for accessory or outfit recommendations, and producing results that satisfy surface-level multimodal relevance but miss the user's true commercial intent.

To address these challenges, we identify three fundamental requirements for an effective e-commerce multimodal search model:

\begin{enumerate}
    \item \textbf{Native e-commerce domain understanding.} The model must internalize rich product knowledge, such as spanning visual attributes, textual descriptions, categorical taxonomies, and cross-product relationships, as well as user behavioral patterns including click sequences, purchase co-occurrence, and preference trajectories. This domain knowledge should be deeply integrated into the model's representations rather than grafted on through external modules.

    \item \textbf{Preservation of general reasoning capabilities.} Domain adaptation through continual pre-training notoriously causes catastrophic forgetting of the base model's language understanding and reasoning abilities. Since the original pre-training corpora of foundation models are typically proprietary and inaccessible, naive fine-tuning on domain data risks degrading the very capabilities that make large language models powerful for complex instruction following and multi-step reasoning.

    \item \textbf{End-to-end product generation with fine granularity.} The model should directly generate product identifiers from multimodal queries in an autoregressive fashion, supporting diverse search intents, such as similar product retrieval, attribute-level editing, style or functionality matching, compound-intent search, and topic-related exploration---within a unified generation framework, while maintaining the fine-grained discrimination necessary to distinguish among billions of candidate products.
\end{enumerate}

In this work, we present \textbf{Pailitao-MMSearch}, a native e-commerce multimodal search foundation model that addresses all three requirements through a cohesive technical framework. Our approach builds upon Qwen3~\citep{qwen3} and Qwen2-VL~\citep{qwen2vl} architecture and introduces three interconnected innovations:

First, we propose \textbf{HybSID} (Hybrid Semantic ID), a novel product tokenization mechanism that represents each product as a compact hybrid of discrete semantic codes and a continuous multimodal embedding. The discrete codes, obtained through residual quantization~\citep{rqvae}, provide coarse-to-fine categorical grouping that enables efficient autoregressive generation, while the continuous embedding preserves fine-grained visual and attribute information that discrete quantization inevitably loses. A shared special token \token{emb\_token} allows the model to generate product embeddings at arbitrary positions, supporting flexible multi-product and interleaved text-product generation.

Second, we design a \textbf{two-stage continual pre-training (CPT)} pipeline that decouples domain knowledge injection from capability preservation. Stage~1 focuses on e-commerce knowledge acquisition through two complementary objectives---product knowledge alignment (understanding individual product attributes and cross-modal correspondences) and user behavior alignment (modeling collaborative signals and sequential patterns). Stage~2 employs On-Policy Distillation (OPD) to mitigate catastrophic forgetting and recover instruction-following capabilities, producing a domain-adapted foundation with restored instruction-following ability for downstream tasks.

Third, we develop a \textbf{hybrid reasoning post-training} pipeline that combines multi-task supervised fine-tuning on diverse e-commerce search scenarios, difficulty-aware reasoning that applies chain-of-thought for complex intents while using direct generation for straightforward queries, reinforcement learning with verifiable rewards grounded in product generation accuracy, and multi-expert OPD that fuses specialized RL expert models with the general-capability base model into a unified final model.
We summarize our main contributions as follows:
\begin{itemize}
    \item We propose HybSID, a hybrid product tokenization scheme that combines discrete semantic IDs with continuous multimodal embeddings, enabling fine-grained product understanding and end-to-end autoregressive product generation within a single language model framework.
    \item We introduce a two-stage continual pre-training strategy that effectively injects e-commerce domain knowledge, including product understanding and user behavior modeling, while mitigating catastrophic forgetting and recovering instruction-following capabilities through OPD.
    \item We develop a hybrid reasoning post-training pipeline featuring difficulty-aware reasoning, reinforcement learning with verifiable product-grounded rewards, and multi-expert knowledge fusion, enabling the model to handle both simple and complex multimodal search intents.
    \item We deploy Pailitao-MMSearch on Taobao's Pailitao platform serving tens of millions of daily active users. Extensive online A/B testing demonstrates significant improvements across key business metrics, including up to +13.61\% GMV and +8.21\% transaction volume, validating the practical effectiveness of our approach.
\end{itemize}

\section{Related Work}
\label{relatedwork}

\subsection{Multimodal Large Language Models}

The emergence of large language models (LLMs)~\citep{gpt4,qwen2.5,qwen3} has catalyzed rapid progress in multimodal understanding. Vision-language models such as LLaVA~\citep{liu2024llava}, Qwen-VL~\citep{qwen2vl,qwen3}, and InternVL~\citep{internvl} extend LLMs with visual perception by aligning image encoders with language decoders through visual instruction tuning. These models demonstrate strong capabilities in general-purpose visual question answering, image captioning, and visual reasoning. However, their training on broad web-scale corpora means they lack the specialized domain knowledge required for e-commerce applications, especially for fine-grained product attribute understanding, commercial taxonomy awareness, and user behavior modeling. Our work addresses this gap by designing domain-adaptive training strategies that inject e-commerce knowledge into VLMs while preserving their general reasoning capabilities.

\subsection{Generative Retrieval with Semantic IDs}
Generative retrieval reformulates the retrieval problem as sequence generation, where a model directly produces document or item identifiers given a query~\citep{dsi,genre}. A critical design choice is how to represent items as token sequences amenable to autoregressive generation. Recent work has explored semantic IDs constructed through hierarchical quantization: TIGER~\citep{tiger} applies Residual-Quantized Variational Autoencoders (RQ-VAE)~\citep{rqvae} to encode item embeddings into multi-level discrete codes, achieving strong performance in recommendation tasks. LC-Rec~\citep{lcrec} further integrates semantic IDs with LLMs for sequential recommendation. While these approaches demonstrate the viability of generative item retrieval, their purely discrete representations sacrifice fine-grained discrimination, which means products that are semantically close but commercially distinct may share identical codes. Our proposed HybSID addresses this limitation by augmenting discrete semantic codes with continuous multimodal embeddings, preserving fine-grained product information within the generative framework.

\subsection{LLMs for E-Commerce Search and Recommendation}
The application of LLMs to search and recommendation has attracted growing interest~\citep{p5,instructrec,productgpt,chen2025onesearch,zhou2025onerecv2technicalreport}. P5~\citep{p5} frames diverse recommendation tasks as natural language generation, unifying rating prediction, sequential recommendation, and explanation generation within a single model. OneRec-V2~\citep{zhou2025onerecv2technicalreport} reformulates recommendation as autoregressive generation over semantic IDs and demonstrates large-scale industrial deployment with high model FLOPs utilization, while OneSearch~\citep{chen2025onesearch} extends the generative paradigm to end-to-end e-commerce search by unifying the traditional multi-stage cascading architecture (retrieval, pre-ranking, ranking) into a single generative framework. Subsequent work has explored instruction-tuned recommendation~\citep{instructrec}, conversational product search~\citep{convprodsearch}, and multimodal product understanding~\citep{productgpt}. In the e-commerce domain, models must handle the unique challenge of operating over massive product catalogs (billions of items) with rich multimodal attributes while serving high-throughput online traffic. \emph{Existing approaches typically rely on external retrieval modules or tool-calling pipelines to bridge the gap between language model generation and product catalog access, introducing system complexity and latency.} Our work instead pursues a native integration approach: by tokenizing the entire product catalog within the model's vocabulary through HybSID and training on e-commerce-specific objectives, Pailitao-MMSearch has achieved end-to-end multimodal product generation for efficient industrial deployment.

\subsection{Continual Pre-Training and Domain Adaptation}
Adapting foundation models to specialized domains through continual pre-training has shown effectiveness across various fields including medicine~\citep{meditron}, finance~\citep{fingpt}, and code generation~\citep{codellama}. A fundamental challenge is catastrophic forgetting: as models acquire domain knowledge, they tend to lose general capabilities acquired during original pre-training~\citep{catastrophicforgetting,luo2025empiricalstudy}. Prior mitigation strategies include replay-based methods that mix domain data with general corpora, regularization-based approaches such as elastic weight consolidation~\citep{ewc}, and progressive training curricula. However, most approaches assume access to the original pre-training data distribution, which is often unavailable for proprietary foundation models. Our two-stage CPT with on-policy distillation provides an alternative: by distilling from the base model's output distribution in Stage~2, we recover general language capabilities without requiring the original training corpus, enabling effective domain adaptation even when the pre-training data is inaccessible.

\section{Method} \label{method}

We present Pailitao-MMSearch, a native e-commerce  multimodal search foundation model that reformulates e-commerce multimodal search as \emph{end-to-end autoregressive product generation}. As illustrated in Figure~\ref{fig:overview}, our approach consists of three main components: (1)~HybSID, a hybrid product tokenization scheme which unifies discrete semantic codes with a continuous multimodal embedding for unified product understanding and generation (\S\ref{sec:hybsid}); (2)~a two-stage continual pre-training pipeline for e-commerce domain knowledge injection and general capability preservation (\S\ref{sec:cpt}); and (3)~a hybrid reasoning post-training pipeline that adapts the model to diverse search intents and tasks (\S\ref{sec:posttraining}).


\begin{figure}[t]
   \begin{center}
\includegraphics[width=0.8\textwidth] {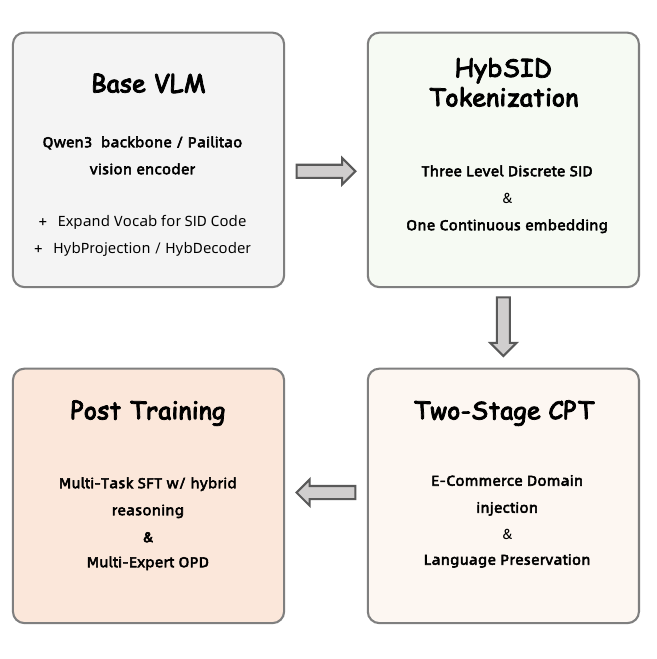}
\end{center}
\vspace{-0.3cm}
    \caption{Overall framework of Pailitao-MMSearch. The system builds upon a vision-language foundation model, introduces HybSID for product tokenization, applies two-stage continual pre-training for domain adaptation, and performs hybrid reasoning post-training for downstream search tasks.}
  \label{fig:overview}
\end{figure}

\subsection{Overview and Problem Formulation} \label{sec:overview}

\paragraph{Problem formulation.}
Let $\mathcal{P} = \{p_1, p_2, \ldots, p_N\}$ denote the product catalog containing $N$ products, where each product $p_i$ is associated with multimodal information including a product image $I_i$, a textual title $t_i$, and structured attributes $a_i$ (e.g., category, brand, material). A user $u$ is characterized by a historical interaction sequence $\mathcal{H}_u = (p_{u,1}, p_{u,2}, \ldots, p_{u,T})$ representing their chronologically ordered browsing, clicking, and purchasing behaviors. A multimodal search query $q = (I_q, t_q)$ consists of an optional query image $I_q$ and optional free-form text $t_q$, where at least one modality is present.

The multimodal search task is formulated as: given a user $u$ with history $\mathcal{H}_u$ and a multimodal query $q$, generate a ranked list of relevant products $\hat{\mathcal{P}} \subset \mathcal{P}$ that satisfy the user's search intent. Pailitao-MMSearch approaches this as an end-to-end autoregressive product generation problem: 、\emph{the model directly generates the target product from the raw multimodal (image and text) queries in an autoregressive fashion, without decomposing them into single-modal sub-queries.} The product is encoded as a discrete-continuous token sequence $z(p)$ (HybSID tokens), conditioned on the multimodal query and user context:

\begin{equation}
    p_\theta\!\left(z \mid q,\mathcal{H}_u\right) \;=\; \prod_{j=1}^{|z|} p_\theta\!\left(z_j \mid z_{<j}, q, \mathcal{H}_u\right).
    \label{eq:ar}
\end{equation}

\begin{table}[t]
\centering
\caption{Notation.}
\label{tab:notation}
\begin{tabular}{ll}
\toprule
Symbol & Meaning \\
\midrule
$q=(I_q,t_q)$ & multimodal query (image, text) \\
$\mathcal{P},\,N$ & product catalog and its size ($\sim\!10^9$) \\
$p_i=(I_i,t_i,a_i)$ & product: image, title, structured attributes \\
$\mathcal{H}_u$ & interaction history of user $u$ \\
$\mathbf{c}=(c_1,c_2,c_3)$ & discrete semantic codes, $c_l\in\{1,\dots,K\}$ \\
$L,\,K$ & codebook levels ($L{=}3$), codebook size ($K{=}8192$) \\
$\mathbf{e}\in\mathbb{R}^{d}$ & continuous product embedding \\
\token{emb\_token} & special token that emits $\mathbf{e}$ \\
$z(p)$ & HybSID token sequence of product $p$ \\
\bottomrule
\end{tabular}
\end{table}

\paragraph{Architecture.}
Pailitao-MMSearch is built on Qwen3 ~\citep{qwen3} with a frozen vision encoder which was trained with tens of billions of Pailitao's visual search data.  We adopt the visual-encoding scheme of Qwen2-VL~\citep{qwen2vl}, maps the reference image $I$ into visual tokens that are concatenated with the tokenized instruction $t$ and consumed by the language model. We extend the LLM backbone in two ways: (i)~the vocabulary is augmented with the HybSID discrete semantic codes and a shared special token \token{emb\_token} for \emph{continuous} product embedding; and (ii)~we introduce an MLP-based \textbf{HybProjection} and \textbf{HybDecoder} that operate on the input and output sides, respectively: the projection aligns the dimensionality of the input-side \emph{continuous} product embedding, while the decoder maps the hidden state of \token{emb\_token} back to a \emph{continuous} product embedding. Table~\ref{tab:notation} summarizes the notation.

\subsection{HybSID: Hybrid Semantic Product Tokenization}
\label{sec:hybsid}

A core challenge in generative product retrieval is representing products as token sequences that a language model can generate autoregressively. Existing Semantic ID (SID) approaches~\citep{tiger,dsi} apply residual quantization to encode a product as a sequence of discrete codes. This enables generative retrieval and yields semantic generalization through the shared codebook, but the information bottleneck of discrete quantization inevitably sacrifices fine-grained discrimination: products that are semantically similar yet commercially distinct may collapse onto identical codes.

To address this, we propose \textbf{HybSID} (Hybrid Semantic ID), which augments the discrete codes $\mathbf{c}$ with a continuous multimodal embedding $\mathbf{e}$, letting the model exploit the generalization of discrete tokenization and the discriminative power of continuous representations at once. The discrete codes provide coarse-to-fine grouping for tractable generation, while the continuous embedding restores the fine-grained detail quantization loses.

\subsubsection{Discrete Semantic Codes via Residual Quantization}

For each product $p_i$, we first extract a dense multimodal representation $\mathbf{v}_i \in \mathbb{R}^d$ using a pre-trained multimodal encoder that jointly encodes the product image, title, and attributes. We then apply Residual Quantization (RQ)~\cite{rqvae} with $L = 3$ levels of quantization:
\begin{equation}
    c_{i,l} = \arg\min_{j \in [K]} \left\| \mathbf{r}_i^{(l-1)} - \mathbf{s}_j^{(l)} \right\|_2, \quad l = 1, 2, 3
\end{equation}
where $\mathbf{s}_j^{(l)} \in \mathcal{C}^{(l)}$ is the $j$-th codeword in the $l$-th level codebook $\mathcal{C}^{(l)} = \{\mathbf{c}_1^{(l)}, \ldots, \mathbf{c}_K^{(l)}\}$ with $K=8192$ entries by following TaoSID2.0-MM-CF~\cite{fu2026forge}, and the residual at each level is:
\begin{equation}
    \mathbf{r}_i^{(0)} = \mathbf{v}_i, \quad \mathbf{r}_i^{(l)} = \mathbf{r}_i^{(l-1)} - \mathbf{s}_j^{(l)}, \quad l = 1, 2, 3
\end{equation}
The coarse-to-fine ordering means $c_1$ groups products into broad semantic clusters and $c_2,c_3$ progressively refine them; the code space $K^{L}=8192^{3}\!\approx\!5.5\times10^{11}$ comfortably covers the billion-scale catalog while keeping each product to only three generated code tokens. These codes are added to the LLM's vocabulary as special SID tokens.

\subsubsection{Continuous Multimodal Embedding and \token{emb\_token}} \label{sec:hybsid-cont}
To complement the discrete codes, each product retains a continuous multimodal embedding $\mathbf{e} \in \mathbb{R}^{d}$ that captures fine-grained product information beyond the resolution of quantized codes. This embedding is produced by a separate encoder from the one used for quantization, and preserves discriminative details such as subtle visual patterns, attribute-level distinctions, and style characteristics.

 We introduce a shared special token \token{emb\_token} into the vocabulary that serves as a \emph{placeholder} for model's embedding input or generation. 

 During the generation, whenever the model emits \token{emb\_token}, its last-layer hidden state $\mathbf{h}$ at that position is projected to the product embedding space. This \token{emb\_token} enables flexible interleaved generation of text and product representations.

\subsubsection{Unified Understanding and Generation}

The complete HybSID representation for product $p_i$ is defined as:
\begin{equation}
    \mathcal{z}(p_i) = \left(c_{1}^i,\; c_2^i,\; c_3^i,\; \mathbf{e}_i\right)
\end{equation}

HybSID operates in two complementary modes:

\textbf{Understanding mode.} When a product appears in the model's input context (e.g., as part of user history or a reference product in the query), its HybSID representation is converted to the LLM's embedding space. The discrete codes are mapped through the standard token embedding lookup, while the continuous embedding is projected via a learned \textbf{HybProjection} module:
\begin{equation}
    \mathbf{x}_i = \left[\mathrm{Emb}(c_{1}^i);\; \mathrm{Emb}(c_{2}^i);\; \mathrm{Emb}(c_{3}^i);\; \mathbf{W}_{\mathrm{proj}} \cdot \mathbf{e}_i + \mathbf{b}_{\mathrm{proj}}\right]
    \label{eq:understanding}
\end{equation}
where $\mathrm{Emb}(\cdot)$ denotes the LLM's token embedding function, $[\cdot;\cdot]$ denotes concatenation along the sequence dimension, and $\mathbf{W}_{\mathrm{proj}} \in \mathbb{R}^{d_{\mathrm{model}} \times d}$ is the projection matrix.

\textbf{Generation mode.} The model autoregressively produces the three discrete SID tokens followed by \token{emb\_token}; its last-layer hidden state $\mathbf{h}$ at that position is passed through a learned \textbf{HybDecoder} to reconstruct the continuous multimodal embedding.
\begin{equation}
    \hat{\mathbf{e}}_i = \mathrm{HybDecoder}(\mathbf{h}_{\mathrm{emb}})
    \label{eq:generation}
\end{equation}
The HybDecoder is a lightweight MLP that maps from the LLM's hidden dimension to the multimodal embedding space. The reconstructed embedding $\hat{\mathbf{e}}_i$ can then be used for fine-grained product retrieval via nearest-neighbor search in the embedding space, complementing the coarse retrieval provided by the discrete SID codes.


\begin{figure}[t]
\begin{center}
\includegraphics[width=0.8\textwidth] {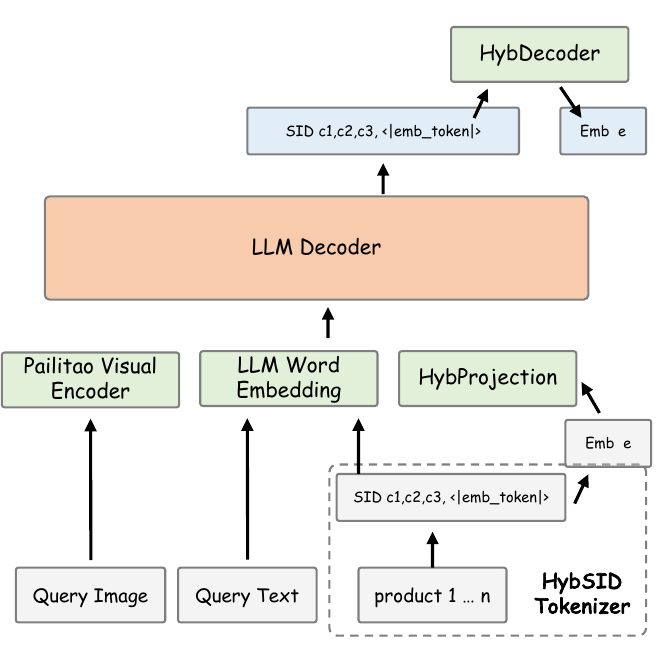}
\end{center}
\vspace{-0.3cm}
\caption{Architecture of HybSID. Each product is represented by three discrete SID tokens obtained through residual quantization and one continuous multimodal embedding. In understanding mode (left), both components are projected into the LLM's embedding space. In generation mode (right), the model autoregressively generates SID tokens and a special \token{emb\_token}, whose hidden state is decoded into the continuous embedding via HybDecoder.}
\label{fig:overview}
\end{figure}
  
The training loss for HybSID generation combines discrete and continuous objectives:
\begin{equation}
    \mathcal{L}_{\mathrm{hyb}} = \mathcal{L}_{\mathrm{CE}}\!\left(\hat{c}^{(1:3)}, c^{(1:3)*}\right) + \lambda \cdot \mathcal{L}_{\mathrm{emb}}\!\left(\hat{\mathbf{e}}, \mathbf{e^{*}}\right)
    \label{eq:hybsid_loss}
\end{equation}
where $\mathcal{L}_{\mathrm{CE}}$ is the standard cross-entropy loss over the discrete SID tokens, $\mathcal{L}_{\mathrm{emb}}$ is a mean squared error loss for the continuous embedding reconstruction, $*$ denotes ground-truth targets, and $\lambda$ is a balancing coefficient.

\subsection{Two-Stage Continual Pre-training} \label{sec:cpt}

\begin{figure}[t]
\begin{center}
\includegraphics[width=0.8\textwidth] {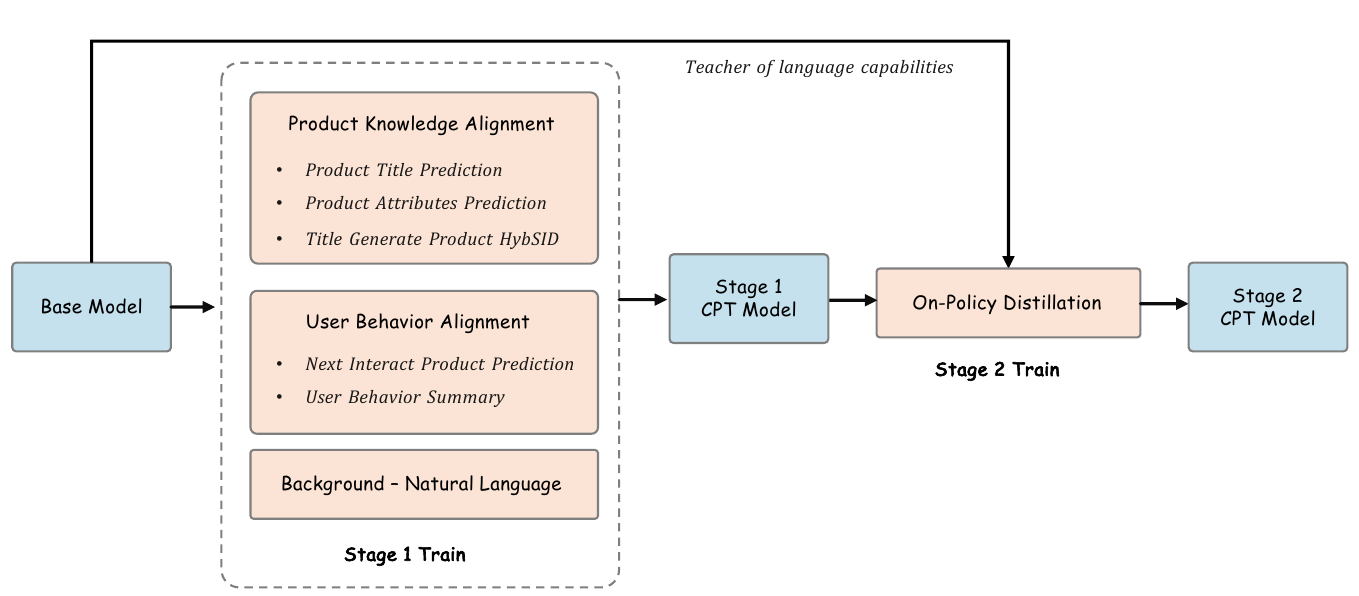}
\end{center}
\vspace{-0.3cm}
    \caption{Two-stage continual pre-training pipeline. Stage~1 injects e-commerce domain knowledge through product knowledge alignment and user behavior alignment tasks, with a background natural language objective. Stage~2 applies On-Policy Distillation (OPD) using the original base LLM as a teacher to recover general language capabilities.}
    \label{fig:cpt}
\end{figure}

Adapting a general-purpose LLM or VLM to the e-commerce domain requires injecting substantial domain knowledge while avoiding catastrophic forgetting of the model's general language understanding and reasoning capabilities. We address this through a carefully designed two-stage continual pre-training (CPT) pipeline, as illustrated in Figure~\ref{fig:cpt}.

\subsubsection{Stage 1: E-Commerce Domain Knowledge Injection}

The first stage focuses on aligning the model with e-commerce domain knowledge through two complementary training tasks, supplemented by a background of natural language task:

\textbf{Product Knowledge Alignment.} This task trains the model to understand individual products at a fine-grained multimodal level. Given a product $p_i$ with its HybSID, the model learns to:
\begin{itemize}
    \item Generate product titles and attribute descriptions from product HybSID;
    \item Reconstruct product HybSID representations from textual and visual descriptions, learning the mapping from natural language product specifications to tokenized product identifiers;
\end{itemize}

\textbf{User Behavior Alignment.} This task captures collaborative signals and sequential patterns from user interaction data. Given a user's historical interaction sequence $\mathcal{H}_u = (p_{u,1}, \ldots, p_{u,T})$, the model learns to:
\begin{itemize}
    \item Predict the next product a user is likely to interact with, given their behavioral history represented as a sequence of HybSID tokens;
    \item Understand product co-occurrence patterns (e.g., users who buy phone cases also buy screen protectors);
    \item Capture preference transitions across sessions and contexts.
\end{itemize}

\textbf{Background Natural Language Task.} To mitigate catastrophic forgetting during Stage~1, we include a small proportion of general-domain conversational data sampled from a distribution similar to the base model's original training data. This serves as a regularizer that prevents the model from drifting too far from its general language capabilities.

\subsubsection{Stage 2: Language Capability Preservation via On-Policy Distillation}

While the background NL task in Stage~1 partially mitigates forgetting, we observe that the model's general language capabilities (instruction following, reasoning, dialogue coherence) still degrade noticeably after extensive domain pre-training. Since the original pre-training corpus of the base model is proprietary and inaccessible, we cannot simply replay it.

We address this through \textbf{On-Policy Distillation (OPD)}, which uses the \emph{original base model} as a teacher to restore the domain-adapted model's language capabilities. Specifically, given a set of general-domain prompts $\mathcal{D}_{\mathrm{gen}}$, we sample responses \emph{on-policy} from the Stage~1 student model $\pi_{\mathrm{s1}}$, and at every generated token align its output distribution to that of the teacher $\pi_{\mathrm{base}}$, while retaining the acquired e-commerce knowledge:

\begin{equation}
    \mathcal{L}_{\mathrm{stage2}} = \mathcal{L}_{\mathrm{task}} + \mu \cdot \mathbb{E}_{\substack{x \sim \mathcal{D}_{\mathrm{gen}} \\ y \sim \pi_{\mathrm{s1}}(\cdot|x)}}\!\left[\sum_{t=1}^{|y|} D_{\mathrm{KL}}\!\left(\pi_{\mathrm{s1}}(\cdot \mid y_{<t}, x) \;\|\; \pi_{\mathrm{base}}(\cdot \mid y_{<t}, x)\right)\right]
    \label{eq:stage2}
\end{equation}
where $\mathcal{L}_{\mathrm{task}}$ continues the domain-specific objectives from Stage~1, the response $y$ is sampled from the student policy $\pi_{\mathrm{s1}}$ (making the distillation on-policy), $D_{\mathrm{KL}}$ is the token-level Kullback-Leibler divergence, and $\mu$ controls the distillation strength.

We verify this preservation in Section~\ref{exp} on general-language benchmarks spanning knowledge (MMLU-Redux, C-Eval), instruction following (IFEval), code (LiveCodeBench-v5), and multilingual understanding (INCLUDE), against a naive single-stage CPT baseline.

\subsection{Post-Training Pipeline}
\label{sec:posttraining}

Building upon the CPT foundation, we design a multi-phase post-training pipeline that progressively enhances the model's capabilities for diverse e-commerce multimodal search tasks.
\begin{figure}[t]
   \begin{center}
\includegraphics[width=0.8\textwidth] {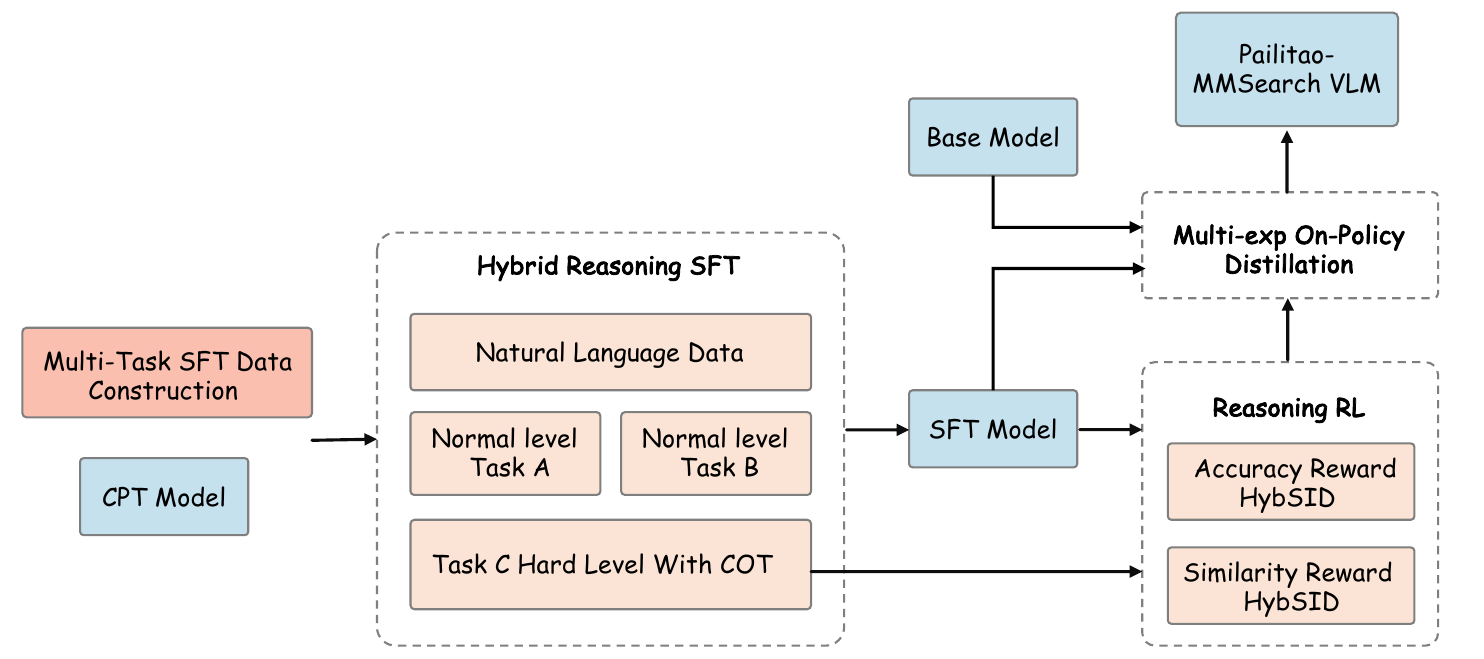}
\end{center}
\vspace{-0.3cm}
    \caption{Post-training pipeline overview. The CPT model undergoes multi-task SFT with difficulty-aware hybrid reasoning, followed by RL enhancement on hard tasks, and finally multi-expert OPD to fuse diverse capabilities into the final model.}
    \label{fig:posttraining}
\end{figure}

\subsubsection{Multi-Task SFT Data Construction}
\label{sec:sft_data}

The key to effective post-training lies in constructing high-quality supervised fine-tuning (SFT) data that covers the diverse search intents encountered in production. We design a systematic data construction pipeline grounded in real user behavior:

\textbf{Step 1: Candidate sample mining.} We extract candidate Query-Doc pairs from user behavioral trajectories across the entire platform. For each predefined search task type (Table~\ref{tab:tasks}), we apply task-specific heuristic rules to identify interaction patterns that match the task semantics. For instance, a user who sequentially clicks on a cup and then purchases a matching cup sleeve triggers the ``matching product'' rule, producing a candidate training pair.

\textbf{Step 2: Instruction generation.} Using the candidate Query-Doc pairs as anchors, we generate task-specific natural language instructions that describe the user's search intent. These instructions are synthesized to be diverse in phrasing while maintaining semantic consistency with the underlying task type.

\textbf{Step 3: Difficulty stratification and quality control.} Candidate samples are stratified by difficulty level (normal, hard) based on the semantic distance between query and target product, the complexity of the search intent, and the number of modalities involved. We apply quality filters to ensure balanced task distribution and appropriate difficulty calibration for the SFT stage.

\begin{table}[t]
\centering
\caption{Multimodal search task types supported by Pailitao-MMSearch. Each task type corresponds to a specific user intent pattern and is associated with dedicated SFT training data.}
\label{tab:tasks}
\small
\begin{tabular}{llp{6cm}}
\toprule
\textbf{Task Type} & \textbf{Modality} & \textbf{Description} \\
\midrule
Similar Product Search & Image (+Text) & Find products visually and functionally similar to a reference image and text instruction if exists \\
Edit \& Modify Product Search & Image + Text & Find products matching the reference image with specific modifications (e.g., color, material) \\
Matching Product Search & Image + Text & Recommend style-matching or functionality-matching items \\
Compound Intent Search & Image + Text & Understand and analyze compound user query and recommend instruct-following products \\
Topic-Related Recommendation & Image (+Text) & Discover topic-related but diverse products for exploration \\
\bottomrule
\end{tabular}
\end{table}

\subsubsection{Hybrid Reasoning: Direct Generation and Chain-of-Thought}
\label{sec:hybrid_reasoning}

Different search scenarios demand different levels of reasoning complexity. A straightforward similar product search may require only direct pattern matching, while a compound intent query (e.g., ``find a formal version of this casual dress in navy blue for a wedding'') requires multi-step reasoning about style transformation, color specification, and occasion appropriateness.

We adopt a \textbf{hybrid reasoning} approach that assigns reasoning strategies based on task difficulty:

\textbf{Direct generation} is applied to normal difficulty samples, where the model directly generates the target product's HybSID given the multimodal query. This mode prioritizes inference efficiency and is suitable for the majority of production traffic.

\textbf{Chain-of-thought (CoT) reasoning} is applied to hard difficulty samples. For these complex scenarios, we construct CoT annotations that decompose the search intent into explicit reasoning steps:
\begin{enumerate}
    \item Identify and describe key attributes of the reference product
    \item Parse the user's modification or search intent
    \item Reason about the target product's expected attributes
    \item Generate the target product's HybSID
\end{enumerate}

The hybrid reasoning SFT is trained with a unified objective where the model learns to produce either direct outputs or CoT-augmented outputs depending on the input's difficulty signal:
\begin{equation}
    \mathcal{L}_{\mathrm{sft}} = -\mathbb{E}_{(x,y) \sim \mathcal{D}_{\mathrm{sft}}} \left[\sum_{l=1}^{|y|} \log P_\theta\!\left(y_l \mid y_{<l}, x\right)\right]
\end{equation}
where $y$ is the target response (either direct HybSID or CoT + HybSID) and $x$ is the multimodal query context.

At inference time, the reasoning mode is specified explicitly in the input instruction according to the online scenario and its latency (RT) constraints.

\subsubsection{Reinforcement Learning with Verifiable Rewards}
\label{sec:rl}

We further optimize product-generation accuracy with reinforcement learning, using Group Relative Policy Optimization (GRPO)~\citep{grpo},  with the verifiable reward defined below.

\noindent\textbf{Verifiable, multi-reference reward.}
The reward is grounded in whether the generated product matches a curated ground-truth set rather than a single logged click, which mitigates exposure bias and reward hacking. For each query we assemble \emph{multiple} ground-truth products $\mathcal{G}(z)$ from expert annotation combined with transaction logs. A generation $o$ is scored as:
\begin{equation}
    R(o) = \alpha_r \cdot \underbrace{\mathbb{1}\!\left[\hat{\mathbf{z}}  \in \mathcal{G}(z)\right]}_{\text{SID Accuracy}} + \beta_r \cdot \underbrace{\mathrm{CosSim}\!\left(\hat{\mathbf{e}},\; \mathbf{e}^{*}\right)}_{\text{Embedding Similarity}}
    \label{eq:reward}
\end{equation}
where $\hat{\mathbf{z}}$ and $\mathcal{G}(z)$ denote the generated and \emph{multiple} ground-truth SID sequences respectively, $\hat{\mathbf{e}}$ and $\mathbf{e}^{*}$ are the generated and ground-truth continuous embeddings, and $\alpha_r$, $\beta_r$ are reward weights.

This reward design has two key advantages: (1)~it is fully automatic and does not require human annotation, enabling scalable RL training over large datasets; (2)~it captures both coarse-level correctness (through SID accuracy) and fine-grained quality (through embedding similarity), providing a richer training signal than binary accuracy alone.

The RL phase produces task-specific \textbf{expert models}---for example, an expert specialized in complex outfit matching or compound intent reasoning---that achieve strong performance on their respective hard task domains.

\subsubsection{Multi-Expert On-Policy Distillation}
\label{sec:multiexpert}

The final stage fuses the capabilities of multiple specialized models into a single unified model through \textbf{Multi-Expert OPD}. We identify three types of expert models:

\begin{enumerate}
    \item \textbf{RL Expert Models}: Multiple task-specific models from the RL phase, each excelling at a particular type of complex search reasoning;
    \item \textbf{Base LLM}: The original language model, which provides strong general natural language understanding and instruction-following capabilities;
    \item \textbf{SFT Model}: The hybrid reasoning SFT model, which serves as the base student model for distillation.
\end{enumerate}

The multi-expert OPD objective trains the student model to selectively absorb capabilities from each expert. Consistent with the on-policy formulation of Stage~2, responses are sampled on-policy from the student and the divergence to each expert is accumulated at every generated token:
\begin{equation}
    \mathcal{L}_{\mathrm{meopd}} = \mathcal{L}_{\mathrm{sft}} + \mathbb{E}_{\substack{x \sim \mathcal{D} \\ y \sim \pi_{\mathrm{student}}(\cdot|x)}}\!\left[\sum_{t=1}^{|y|} \sum_{k=1}^{M} w_k \cdot D_{\mathrm{KL}}\!\left(\pi_{\mathrm{student}}(\cdot \mid y_{<t}, x) \;\|\; \pi_k(\cdot \mid y_{<t}, x)\right)\right]
    \label{eq:meopd}
\end{equation}
where $M$ is the number of expert models, $\pi_k$ represents the $k$-th expert model's policy, $w_k$ is the distillation weight for expert $k$, the response $y$ is sampled on-policy from the student, and the SFT loss $\mathcal{L}_{\mathrm{sft}}$ ensures the student retains its task-solving capabilities.

The final model inherits the domain-specialized reasoning of the RL experts, the general language capabilities of the base LLM, and the multi-task versatility of the SFT model, producing a comprehensive e-commerce multimodal search foundation model ready for production deployment.
\section{Experiments} \label{exp}

We evaluate Pailitao-MMSearch through large-scale online A/B testing on Taobao's Pailitao platform and comprehensive offline studies covering e-commerce domain understanding, general-language capability preservation, and multimodal intent following. Additionally, we conduct component ablations and qualitative case analyses. Specifically, our experiments aim to answer the following questions: (1)~Does Pailitao-MMSearch improve key business metrics under real production traffic? (2)~Does the two-stage CPT pipeline inject product and user-behavior knowledge while preserving general-language and instruction-following capabilities? (3)~How effectively does the post-trained model handle diverse multimodal search intents? (4)~How do key design choices affect model performance?

\subsection{Experimental Setup}

\textbf{Platform and Scale}:
We deploy and evaluate Pailitao-MMSearch in production for generative retrieval and ranking on Pailitao, Taobao's multimodal product search platform serving tens of millions of daily active users. The platform supports diverse search modalities including image-based visual search, image-text combined search, and free-form multimodal queries. The product catalog encompasses billions of active listings spanning all major e-commerce categories (\textit{e.g.} fashion, electronics, home goods, beauty, etc.).

\textbf{Base Model and Training Configuration}:
Pailitao-MMSearch uses Qwen3~\citep{qwen3} as its language backbone, paired with a frozen vision encoder. HybSID represents each product using three residual-quantized semantic codes together with a continuous multimodal embedding. A ground-truth three-level SID bucket contains approximately 24 candidate products on average, motivating the continuous component for fine-grained discrimination within each bucket. Key training hyperparameters are summarized in Table~\ref{tab:hyperparams}.

\begin{table}[t]
\centering
\caption{Training hyperparameters for each stage of Pailitao-MMSearch. For Stage~2, the sequence-length entry reports the maximum prompt/response lengths.}
\label{tab:hyperparams}
\small
\begin{tabular}{lccc}
\toprule
\textbf{Hyperparameter} & \textbf{CPT Stage 1} & \textbf{CPT Stage 2} & \textbf{Post-Train SFT} \\
\midrule
Data Scale(num. of samples) & 0.8B & 2M & 30M \\
Learning Rate & 8e-5 & 3e-5 & 5e-6 \\
Global Batch Size & 64 & 14 & 384 \\
Sequence Length & 4096 & 1024 / 4096 & 4096 \\
Warmup Ratio & 0.003 & 0 & 0.05 \\
Weight Decay & 0 & 0.01 & 0.01 \\
\bottomrule
\end{tabular}
\end{table}

The reported Stage~2 checkpoint is obtained by equal-weight interpolation between the Stage~1 checkpoint and the OPD endpoint, i.e., $\theta_{\mathrm{s2}}=0.5\theta_{\mathrm{s1}}+0.5\theta_{\mathrm{opd}}$.

\textbf{Evaluation Metrics}:
We evaluate Pailitao-MMSearch using complementary online and offline protocols.

\textbf{Online evaluation.} For deployed generative retrieval and ranking, we report relative lifts over the production baseline using standard e-commerce metrics. \textbf{GMV (Gross Merchandise Volume)} measures the total transaction value. \textbf{Transaction Volume} is the number of completed purchase transactions, while \textbf{Transaction UV} is the number of unique users completing transactions. \textbf{IPV (Item Page Views)} measures the number of product detail page visits, while \textbf{IPV\_UV} counts the unique users visiting product detail pages. \textbf{CTR (Click-Through Rate)} is the ratio of clicks to impressions, and \textbf{CVR (Conversion Rate)} is the ratio of purchases to clicks.

\textbf{Offline evaluation.} Our offline evaluation covers four capability groups:
\begin{itemize}
    \item \textbf{Product understanding.} SID translation is evaluated using BLEU-4, ROUGE-L, exact match, and attribute exact match. Multimodal SID recognition is evaluated using SID-1 accuracy, SID-1/2 accuracy, Full SID Match, Oracle Embedding Cosine, and End-to-end Retrieval@1/@10.
    \item \textbf{User-sequence modeling.} We report SID-1 accuracy, SID-1/2 accuracy, and Full SID Match.
    \item \textbf{General-language capability.} We report normalized benchmark scores spanning knowledge, reasoning, mathematics, code generation, and multilingual understanding.
    \item \textbf{Post-training multimodal search.} After SFT and RL, we report Recall@10 and NDCG@10 for retrieval quality, together with judge-based (Strict Instruction Following) SIF@10 and (Loose Instruction Following) LIF@10 for strict and loose instruction following.
\end{itemize}

\subsection{Downstream Application and Online Performance}
We deploy Pailitao-MMSearch in two primary application scenarios: generative retrieval and generative ranking of Pailitao's ``PeiWoGuang'' floor, which aims to find similar items given user image queries.

\subsubsection{Generative Retrieval}
In this application, given an image query, Pailitao-MMSearch autoregressively generates product HybSIDs and retrieves products through SID-based routing and embedding matching. Unlike traditional retrieval methods based on direct query--product matching, our approach formulates candidate generation as HybSID generation. The resulting candidates complement existing retrieval channels.

\begin{table}[t]
\centering
\caption{Online A/B test results for generative retrieval. Results are reported for both the Pailitao overall platform and the ``PeiWoGuang'' floor specifically.}
\label{tab:recall}
\begin{tabular}{lcc}
\toprule
\textbf{Metric} & \textbf{PeiWoGuang Floor} & \textbf{Pailitao Overall} \\
\midrule
GMV & +3.67\% & +1.02\% \\
Transaction Volume & +3.74\% & +0.53\% \\
Transaction UV & +2.51\% & -- \\
IPV & +3.28\% & +0.15\% \\
CTR & +0.11pt & +0.01pt \\
CVR & +0.00pt & +0.01pt \\
\bottomrule
\end{tabular}
\end{table}


Table~\ref{tab:recall} presents the results. The generative retrieval model achieves \textbf{+1.02\% GMV} at the overall Pailitao scenario level and \textbf{+3.67\% GMV} on the ``PeiWoGuang'' floor. The particularly strong PeiWoGuang-floor improvements (+3.74\% transaction volume, +3.28\% IPV) indicate that the model excels at generating novel, relevant products that users would not have discovered through traditional retrieval pipelines. The +2.51\% transaction UV improvement on the ``PeiWoGuang'' floor demonstrates that generative recall attracts new converting users, expanding the platform's commercial reach. 

\subsubsection{Generative Ranking}
In this application, the CPT model is further fine-tuned into a ranking model on the ``PeiWoGuang'' floor. The model takes the user's image query and a set of candidate products as input, and generates a ranked ordering by predicting the click-through rate (CTR) probability.

\begin{table}[t]
\centering
\caption{Online A/B test results for generative ranking on the ``PeiWoGuang'' floor. All improvements are relative to the production baseline.}
\label{tab:reranking}
\begin{tabular}{lcc}
\toprule
\textbf{Metric} & \textbf{PeiWoGuang Floor} & \textbf{Pailitao Overall} \\
\midrule
IPV & +3.52\% & +0.16\% \\
IPV\_UV & +2.11\% & +0.47\% \\
Transaction Volume & +4.47\% & +0.41\% \\
GMV & +9.94\% & +0.25\% \\
CTR & +0.11pt & +0.01pt \\
CVR & +0.01pt & +0.01pt \\
\bottomrule
\end{tabular}
\end{table}

As shown in Table~\ref{tab:reranking}, the generative ranking model achieves substantial improvements on the ``PeiWoGuang'' floor: \textbf{+9.94\% GMV} and \textbf{+4.47\% transaction volume}, demonstrating that the e-commerce domain knowledge acquired through CPT translates directly into superior ranking quality. The improvement in IPV\_UV (+0.47\%) at the overall Pailitao scenario level indicates that better recommendation quality attracts more engaged users. The CTR improvement (+0.11pt) suggests enhanced result relevance, while the CVR lift indicates improved purchase intent matching. Combining the two deployments, Pailitao-MMSearch delivers cumulative gains of \textbf{+13.61\% GMV} and \textbf{+8.21\% transaction volume} on the ``PeiWoGuang'' floor.

\begin{figure}[t]
\begin{center}
\includegraphics[width=0.8\textwidth] {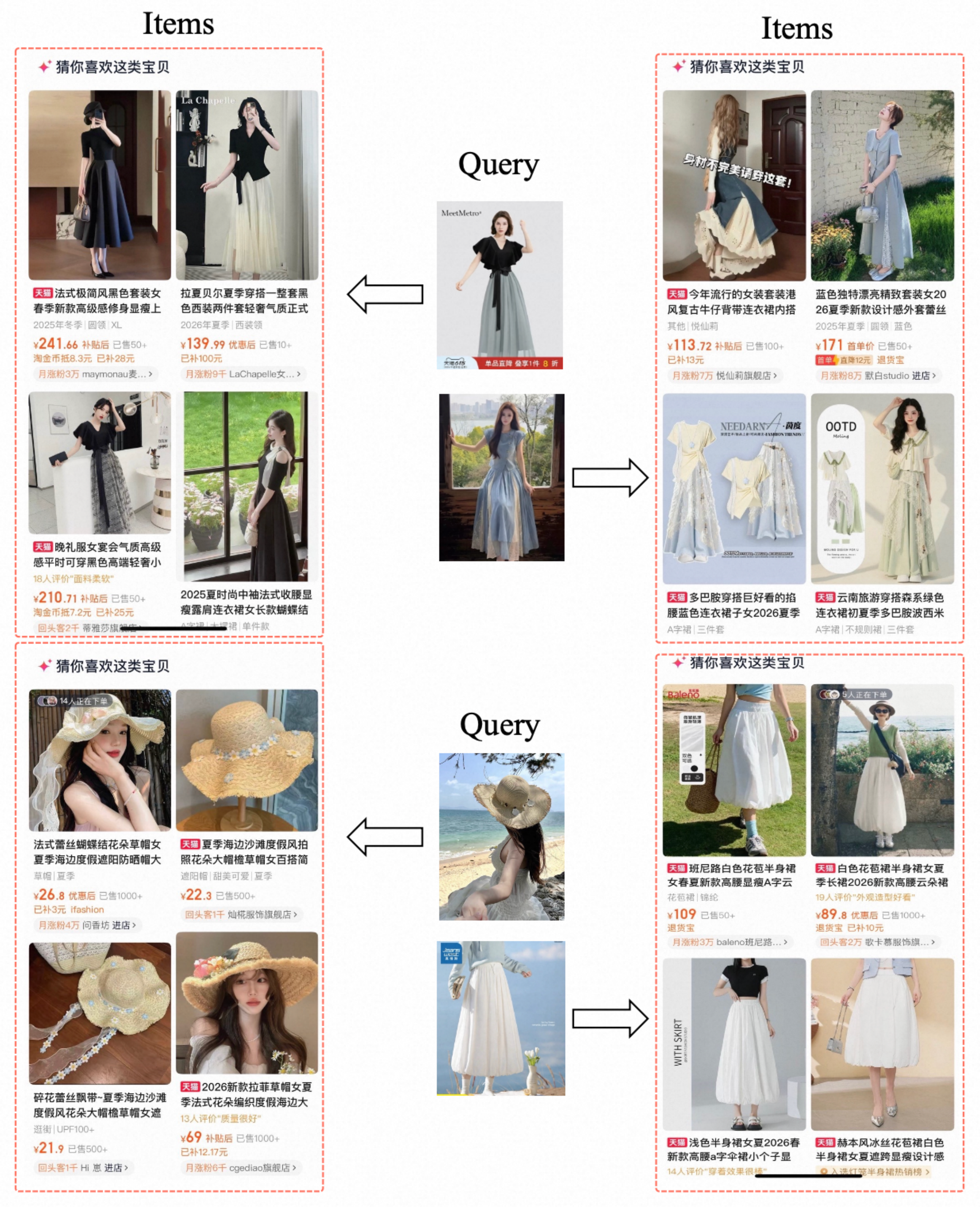}
\end{center}
\vspace{-0.3cm}
\caption{Real-world examples of our generative retrieval and ranking pipeline based on Pailitao-MMSearch in ``PeiWoGuang'' of Pailitao at Taobao app.}
\label{fig:peiwoguang_cases}
\end{figure}

We show some real-world examples of this generative retrieval and ranking pipeline in ``PeiWoGuang'' of Pailitao at Taobao app. In this case, a user upload images and the system aims to recommend style-similar or functionality-similar products. The results are given in Figure~\ref{fig:peiwoguang_cases}.

\subsection{E-commerce Domain Understanding and General Language Abilities}

To verify that our continual pre-training (CPT) effectively injects e-commerce domain knowledge while preserving general language capabilities, we design a comprehensive offline evaluation suite covering two dimensions: (1)~e-commerce domain understanding tasks that assess the model's mastery of product knowledge encoded through HybSID, and (2)~general language ability benchmarks that measure whether the model retains broad reasoning and generation capabilities after domain-specific training.

\subsubsection{E-commerce Domain Understanding}
We evaluate the model's e-commerce knowledge through three categories of tasks that probe different aspects of product understanding. We further evaluate the coarse-to-fine retrieval mechanism enabled by HybSID. The generated SID-1/2 prefix routes a query to a semantically consistent candidate bucket, while the generated continuous embedding ranks products within that bucket. End-to-end Retrieval@K measures the product hit rate of this cascade, requiring both a correct SID-1/2 prediction and successful embedding-based retrieval.

\textbf{SID Translation Tasks.} These tasks assess whether the model has internalized the semantic mapping from structured product identifiers to natural language descriptions:
\begin{itemize}
    \item \textbf{SID$\rightarrow$Caption} (\texttt{sid\_translation\_caption}): Given a product SID, the model generates a detailed natural language description of the product, including its category, appearance, material, and key attributes.
    \item \textbf{SID$\rightarrow$Attributes} (\texttt{sid\_translation\_pv}): Given a product SID, the model generates a structured attribute list (property-value pairs) characterizing the product.
\end{itemize}

\textbf{Multimodal SID Recognition Tasks.} This task evaluate the inverse mapping---whether the model can identify the correct product given various forms of multimodal product information like caption, attributes and multimodal emebddings.

\textbf{User Sequence Prediction.} This task evaluates whether the model has learned meaningful user behavior patterns from the behavior-aligned pre-training. Given a sequence of historically interacted product SIDs, the model predicts the SID of the next product the user is likely to engage with, reflecting its understanding of user preference dynamics and product transition patterns.

\begin{table}[t]
\centering
\caption{E-commerce domain understanding evaluation results. SID-1 and SID-1/2 report prefix-level matching over the hierarchical discrete SID codes, while Full SID Match requires all three discrete SID levels to match exactly. Oracle Embedding Cosine is computed only on examples with an exact three-level SID match. End-to-end Retrieval@K requires a correct SID-1/2 prediction followed by embedding-based product retrieval within the corresponding SID-1/2 bucket. Higher is better for all metrics.}
\label{tab:ecom_understanding}
\small
\begin{tabular}{lcc}
\toprule
\multicolumn{3}{c}{\textbf{(a) SID Translation}} \\
\midrule
\textbf{Metric} & \textbf{SID$\rightarrow$Caption} & \textbf{SID$\rightarrow$Attributes} \\
\midrule
BLEU-4 & 0.1955 & 0.0738 \\
ROUGE-L & 0.4196 & -- \\
Exact / Attr Exact & 0.53\% & 50.52\% \\
\midrule
\multicolumn{3}{c}{\textbf{(b) Multimodal SID Recognition}} \\
\midrule
\textbf{Metric} & \multicolumn{2}{c}{\textbf{Value}} \\
\midrule
SID-1 Acc & \multicolumn{2}{c}{74.78\%} \\
SID-1/2 Acc & \multicolumn{2}{c}{26.95\%} \\
Full SID Match & \multicolumn{2}{c}{2.10\%} \\
Oracle Embedding Cosine & \multicolumn{2}{c}{0.7470} \\
End-to-end Retrieval@1 & \multicolumn{2}{c}{2.10\%} \\
End-to-end Retrieval@10 & \multicolumn{2}{c}{4.43\%} \\
\midrule
\multicolumn{3}{c}{\textbf{(c) User Sequence Prediction}} \\
\midrule
\textbf{Metric} & \multicolumn{2}{c}{\textbf{Value}} \\
\midrule
SID-1 Acc & \multicolumn{2}{c}{34.25\%} \\
SID-1/2 Acc & \multicolumn{2}{c}{12.34\%} \\
Full SID Match & \multicolumn{2}{c}{3.38\%} \\
\bottomrule
\end{tabular}
\end{table}

The results in Table~\ref{tab:ecom_understanding} demonstrate that after CPT, the model acquires strong e-commerce domain understanding. The SID translation tasks confirm that the model can decode product identifiers into rich semantic descriptions, indicating that HybSID tokens are not merely memorized but are grounded in meaningful product representations. The SID recognition task validates the reverse capability---the model can map multimodal product information back to precise identifiers, which is critical for generative recall. Compared with a stricter SID-1/2/3+embedding cascade, which obtains only 1.08\%/1.55\% at @1/@10, the SID-1/2+embedding design improves end-to-end retrieval to 2.10\%/4.43\%, despite searching a substantially larger candidate bucket. User sequence prediction is reserved for behavior-alignment evaluation.

\subsubsection{General Language Abilities}

A known risk of domain-specific continual pre-training is catastrophic forgetting of general capabilities. We evaluate whether our On-Policy Distillation (OPD) strategy in Stage~2 CPT effectively mitigates this issue by benchmarking the model on diverse general-purpose language tasks.

\begin{table}[t]
\centering
\caption{General language ability evaluation across diverse domains. We compare Qwen3-0.6B, the Stage~1 CPT model (before OPD recovery), and the Stage~2 CPT model obtained by equal-weight interpolation ($\alpha=0.50$) between Stage~1 and its OPD endpoint trained with a learning rate of $3\times10^{-5}$. Base-model scores are taken from the official Qwen3 report under the non-thinking setting; all scores are normalized to [0, 100].}
\label{tab:general_language}
\small
{\setlength{\tabcolsep}{3pt}
\renewcommand{\arraystretch}{1.08}
\begin{tabular}{@{}llccc@{}}
\toprule
\textbf{Field} & \textbf{Benchmark} & \textbf{Qwen3-0.6B} & \textbf{\shortstack{After Stage~1\\CPT}} & \textbf{\shortstack{After Stage~2\\CPT (Ours)}} \\
\midrule
\multirow{2}{*}{\textit{General}} & MMLU-Redux & 44.6 & 39.21 & 38.63 \\
 & C-Eval & 42.6 & 35.81 & 36.92 \\
\midrule
\textit{Alignment} & IFEval (strict prompt) & 54.5 & 39.37 & 47.32 \\
\midrule
\textit{Math \& Text Reasoning} & MATH-500 & 55.2 & 16.20 & 32.60 \\
\midrule
\textit{Agent \& Coding} & LiveCodeBench v5 & 3.6 & 2.41 & 2.41 \\
\midrule
\textit{Multilingual} & INCLUDE & 34.4 & 29.47 & 29.01 \\
\bottomrule
\end{tabular}
}
\end{table}

As shown in Table~\ref{tab:general_language}, the Stage~1 CPT model exhibits notable degradation on several general language benchmarks compared to the base model, confirming that aggressive domain-specific training risks capability loss. Relative to Stage~1, the interpolated Stage~2 model improves mathematical reasoning and instruction following: MATH-500 increases from 16.20 to 32.60 (+16.40 points), while IFEval strict-prompt accuracy increases from 39.37 to 47.32 (+7.95 points). MMLU-Redux, C-Eval, and INCLUDE reach 38.63, 36.92, and 29.01, respectively, while LiveCodeBench remains unchanged at 2.41 under the same evaluation protocol. During model selection, this interpolation produced point estimates no lower than a matched LR1e-5 OPD reference on all five displayed non-code benchmarks while preserving the same LiveCodeBench score. Paired analysis identifies a significant recovery only on INCLUDE; the incremental differences on MATH-500, IFEval, MMLU-Redux, and C-Eval are not statistically significant.

\subsection{Promising Instruct-Following Multi-Modal Search}

Beyond offline metrics, we demonstrate the practical potential of Pailitao-MMSearch as an instruct-following multimodal search system. Starting from the CPT base model, we conduct downstream post-training (SFT + RL) to build a system that accepts user-provided images combined with free-form textual instructions, enabling diverse search intents including similar product search, matching product search, and compound intent search. The instruct-following multi-modal search function in our scenario are not online yet, and we will update the online performance in latter arXiv versions when the A/B test results are collected.

The instruct-following multimodal search system takes as input a user-provided reference image $I$ and a textual instruction $T$, and generates a ranked list of product HybSIDs that satisfy the combined multimodal intent. We categorize supported search intents as follows:
\begin{itemize}
    \item \textbf{Similar Product Search}: The user provides a product image and seeks visually and functionally similar products (e.g., ``find similar styles'').
    \item \textbf{Matching Product Search}: The user provides an item image and seeks style-matching or functionality-matching items (e.g., ``matching pants for this top'').
\end{itemize}

\begin{figure}[t]
\begin{center}
\includegraphics[width=1\textwidth] {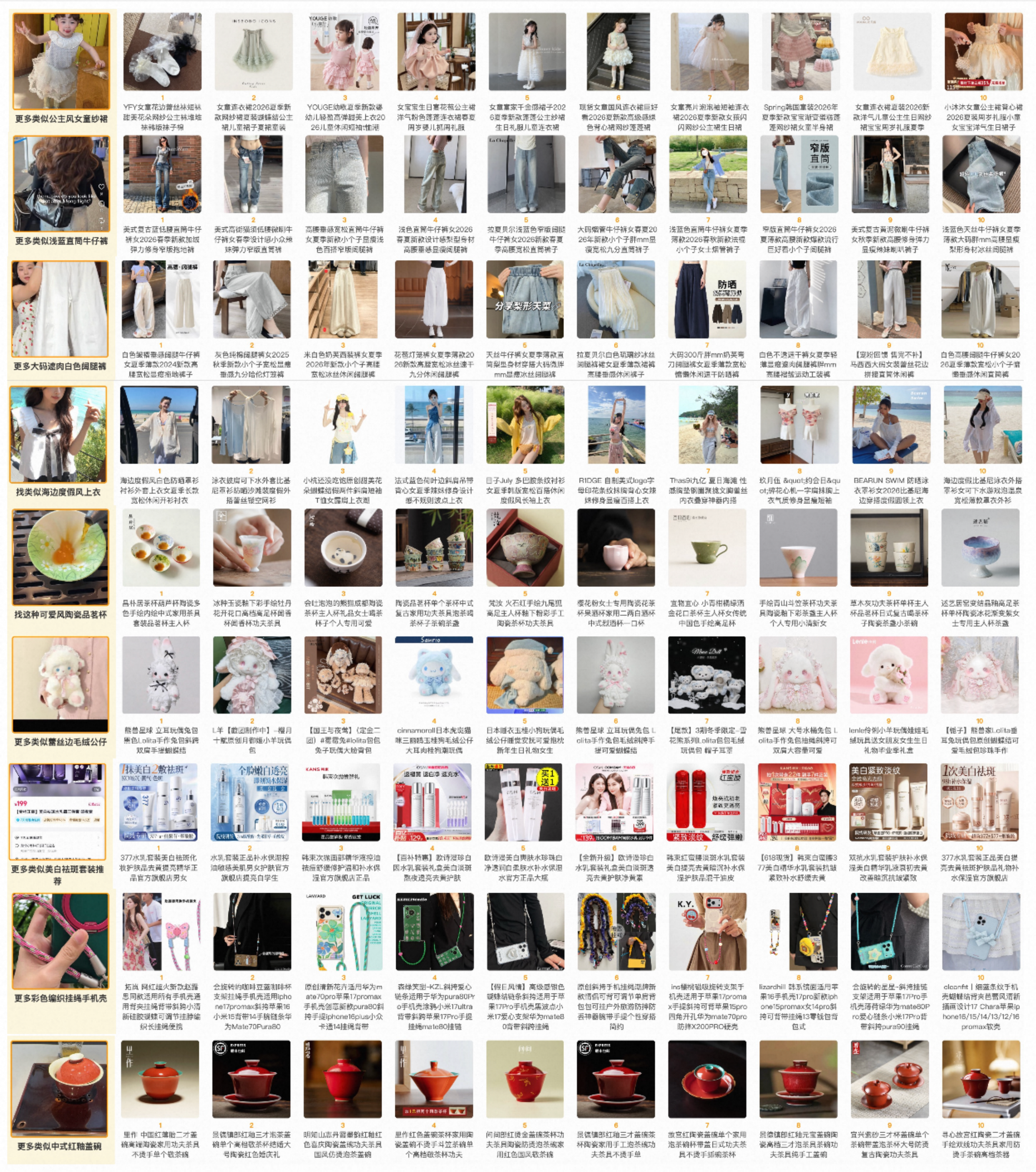}
\end{center}
\vspace{-0.3cm}
\caption{Similar Product Search: User uploads query image and free text, and the model generates similar products following user multi-modal intentions. The first row indicates the user image query and free text, while the second to eleventh rows are top10 generated items by Pailitao-MMSearch model.}
\label{fig:similar_cases}
\end{figure}

\begin{figure}[t]
\begin{center}
\includegraphics[width=1\textwidth] {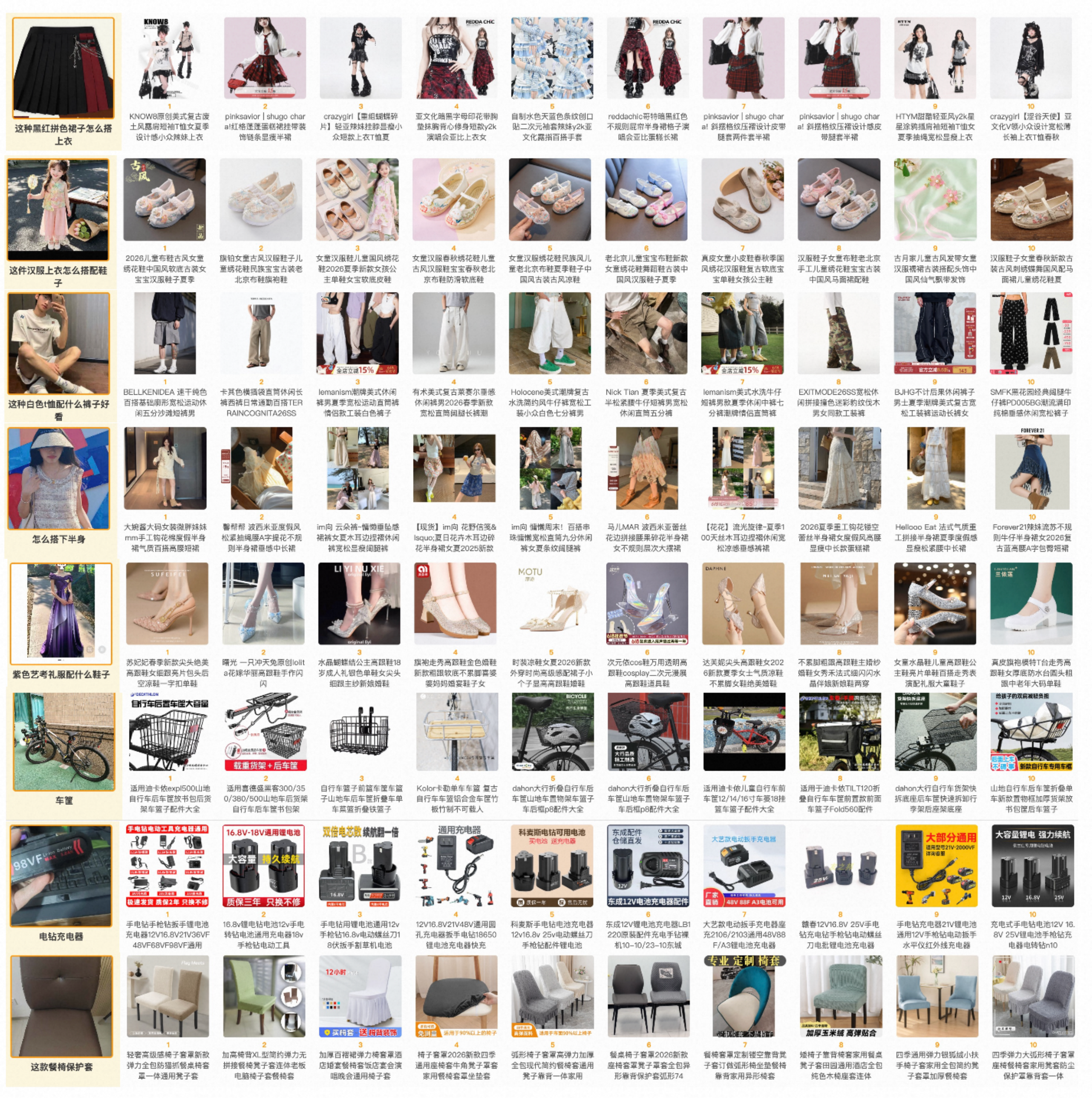}
\end{center}
\vspace{-0.3cm}
\caption{Matching Product Search: User uploads query image and free text, and the model generates matching products following user multi-modal intentions. The first row indicates the user image query and free text, while the second to eleventh rows are top10 generated items by Pailitao-MMSearch model.}
\label{fig:matching_cases}
\end{figure}

\begin{figure}[t]
\begin{center}
\includegraphics[width=1\textwidth] {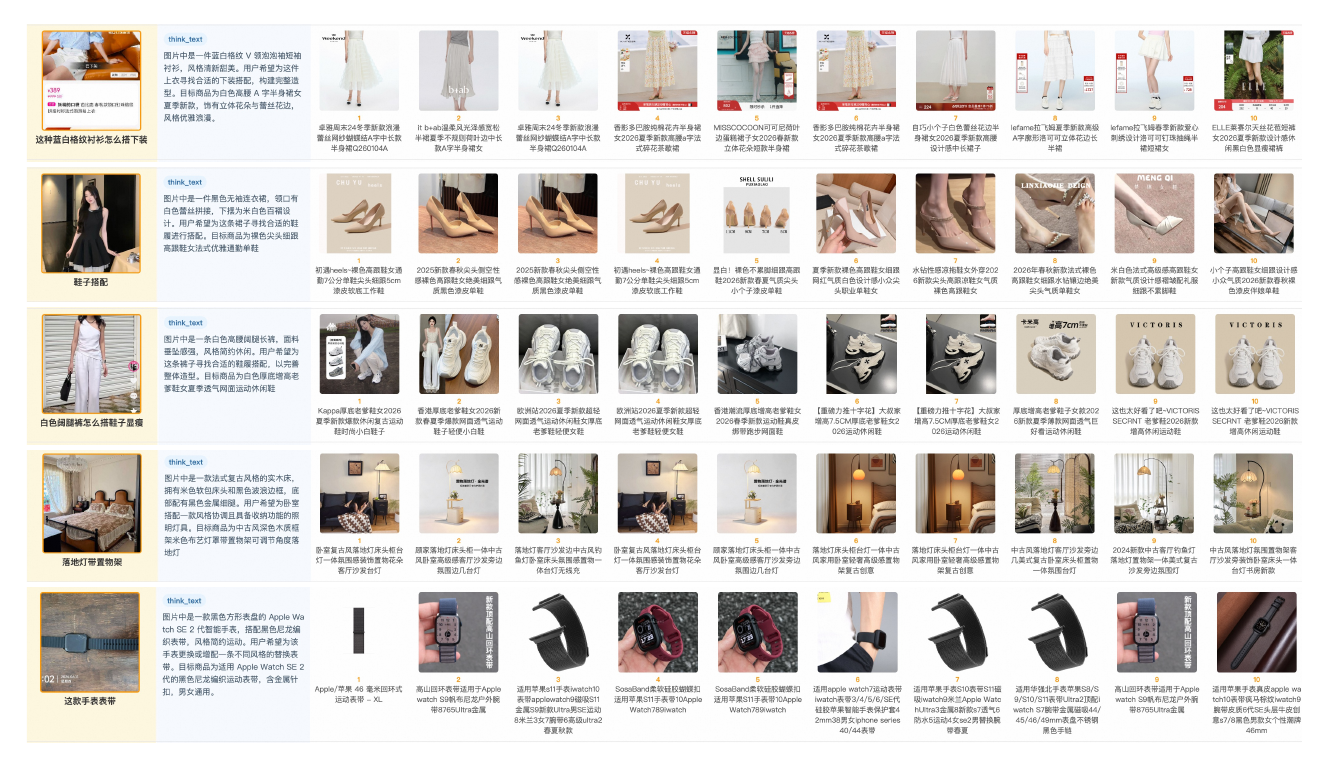}
\end{center}
\caption{Matching Product Search with think mode: User uploads query image and free text, and the model generates thinking text and matching products following user multi-modal intentions. The first row indicates the user image query and free text, while the second to eleventh rows are top10 generated items by Pailitao-MMSearch model.}
\label{fig:matching_think_cases}
\end{figure}

\subsubsection{Quantitative Results}

\begin{table}[t]
\centering
\caption{Instruct-following multimodal search results across different intent categories. We report Recall@K and NDCG@K on held-out test sets for each search intent type. We also use Gemini3.1-pro as judge for Strict Instruction Following Rate (SIF@$k$) and Loose Instruction Following Rate (LIF@$k$). 
}
\label{tab:instruct_search}
\small
\begin{tabular}{lcccc}
\toprule
\textbf{Search Intent} & \textbf{Recall@10} & \textbf{NDCG@10} & \textbf{SIF@10} & \textbf{LIF@10} \\
\midrule
Similar Product Search & 0.8457 & 0.8337 & 43.3\% & 81.6\% \\
Matching Product Search & 0.8544 & 0.8146 & 55.0\% & 80.1\% \\
Matching Product Search(w/ COT) & 0.8356 & 0.8326 & 58.6\% & 82.6\% \\
\midrule
\textbf{Overall} & 0.8452 & 0.8269 & 52.3\% & 81.43\% \\
\bottomrule
\end{tabular}
\end{table}

Table~\ref{tab:instruct_search} presents the quantitative evaluation results. The model demonstrates strong performance across all intent categories, with particularly notable results on similar product search where the HybSID's continuous embedding component enables fine-grained visual similarity matching. Matching Product Search benefits from user behavior alignment knowledge (co-purchase patterns). 

\subsubsection{Qualitative Analysis}
\label{sec:qualitative}

We present representative examples from the deployed system to illustrate the model's multimodal search capabilities across different task types.


\textbf{Similar Product Search.} Given a reference product image, the model generates HybSIDs of visually and functionally similar products, as shown in Figure~\ref{fig:similar_cases}. The continuous embedding component of HybSID enables the model to capture subtle style differences that discrete codes alone would miss, such as distinguishing between Princess-style children's dress and general children's dress that share the same categorical structure.


\textbf{Matching Product Search.} For fashion-oriented queries where users seek complementary items, the model leverages user behavior alignment knowledge (product co-purchase patterns) combined with visual style understanding to generate aesthetically and functionally coherent outfit recommendations. The results in Figure~\ref{fig:matching_cases} show the model's performance on matching product search. We clearly see that in fashion cases, the model can learn different fashion styles and user's multi-modal intentions. Meanwhile, for some functional accessories, the model can also understand the query's characteristic and functionality, and then recommend complementary products matching the user's multi-modal intentions. When enabling think mode, the results are given in Figure~\ref{fig:matching_think_cases}. It is clear that the model has better capability of understand attributes, brand and other implicit information in query images, and makes appropriate products.  


\subsection{Ablation Studies on Model Components}
To understand the contribution of each component, we conduct ablation experiments comparing against the following variants:

\begin{table}[t]
\centering
\caption{Ablation studies across continual pre-training and post-training. Panel (a) reports the e-commerce metrics defined in Table~\ref{tab:ecom_understanding}. For end-to-end retrieval, the full model uses SID-1/2+embedding, while the variant without discrete SIDs performs global embedding-only retrieval. Panel (b) reserves downstream search metrics for pending component studies.}
\label{tab:ablation}
\small
{\setlength{\tabcolsep}{3pt}
\begin{tabular}{@{}lcccccc@{}}
\toprule
\multicolumn{7}{c}{\textbf{(a) HybSID Representation}} \\
\midrule
\textbf{Model Variant} & \textbf{\shortstack{Caption\\BLEU-4}} & \textbf{\shortstack{Attr\\Exact}} & \textbf{\shortstack{SID-1\\Acc}} & \textbf{\shortstack{SID-1/2\\Acc}} & \textbf{\shortstack{Full SID\\Match}} & \textbf{\shortstack{E2E Retrieval\\@1 / @10}} \\
\midrule
Pailitao-MMSearch (Full) & 0.1955 & 50.52\% & 74.78\% & 26.95\% & 2.10\% & 2.10\% / 4.43\% \\
\quad w/o Continuous MM Embedding & 0.1212 & 39.70\% & 75.48\% & 27.05\% & 2.15\% & -- \\
\quad w/o Discrete SID & -- & -- & -- & -- & -- & 1.80\% / 4.08\% \\
\bottomrule
\end{tabular}
}

\end{table}


\textbf{(a) HybSID component ablations.} Under the matched CPT setting, removing the continuous multimodal embedding reduces SID-to-caption BLEU-4 from 0.1955 to 0.1212 and attribute exact match from 50.52\% to 39.70\%, while the three discrete SID matching metrics remain comparable. This variant produces no continuous embedding, so embedding-based end-to-end retrieval is not applicable. The complementary variant without discrete SIDs performs global embedding-only retrieval, reaching 1.80\% and 4.08\% at ranks 1 and 10, respectively.

\textbf{(b) Other components.} Results for hybrid reasoning, RL enhancement, and multi-expert OPD will be updated in later arXiv versions.


\section{Conclusion}
We present Pailitao-MMSearch, one native e-commerce multimodal search foundation model that addresses the fundamental limitations of both single-modal specialist systems and general-purpose vision-language models in e-commerce product search. Our approach introduces three interconnected innovations: HybSID, a hybrid product tokenization scheme that unifies discrete semantic codes with continuous multimodal embeddings for fine-grained product understanding and generation; a two-stage continual pre-training pipeline that injects e-commerce domain knowledge while preserving general language capabilities through on-policy distillation; and a hybrid reasoning post-training framework combining difficulty-aware chain-of-thought reasoning, reinforcement learning with verifiable product-grounded rewards, and multi-expert knowledge fusion.
Deployed on Taobao's Pailitao platform serving hundreds of millions of users, Pailitao-MMSearch achieves substantial and statistically significant improvements across key business metrics. Its deployment in more business cases are on-going and we will update the results in later versions. 

Through our real-world case studies, we have also identified several limitations of the current approach. A non-trivial portion of user queries in practice are highly complex multimodal inputs: images may be captured from unusual angles or suffer from low resolution, and are sometimes further edited by users with additional annotations or markings; meanwhile, the accompanying free-form text queries frequently contain typographical errors, misspellings, or expressions that deviate from standard linguistic conventions. For such challenging cases, the current model still falls short of producing satisfactory results. Our analytical experiments reveal that the primary bottleneck lies in the model's insufficient reasoning capabilities when confronted with these noisy and ambiguous multimodal signals. \textbf{\textit{In the next version of Pailitao-MMSearch, we plan to prioritize comprehensive enhancements to the model's multimodal reasoning abilities to better handle these complex real-world scenarios.}}

Beyond these immediate extensions, we would like to highlight a broader motivation underlying this work. Agentic search systems where autonomous agents orchestrate multi-step retrieval, reasoning, and tool use represent a promising frontier for e-commerce search. \emph{\textbf{However, we argue that the critical bottleneck for such systems now is not the orchestration framework design, but rather the availability of an applicable foundation model with genuine e-commerce domain understanding. An agent equipped with deep product knowledge and user intent comprehension will naturally make more accurate decisions in tool invocation, query reformulation, and result synthesis.}} Building this domain-native foundation is essential infrastructure work that enables, rather than competes with, agentic architectures. In subsequent work, we will also present our agentic search system built upon Pailitao-MMSearch, offering our perspectives and solutions to key challenges in agentic e-commerce search systems.

\section{Contributors}
\noindent\textbf{Core Contributors}: Xiaohan Ye, Xu Chen, Zihan Gong, Jian Ding

\noindent\textbf{Algorithm Contributors}: Lianyu Du, Baicheng Chen, Yunmeng Shu, Jingqian Zhao, Shengxin Nie, Chen Ju, Zhaoyang Li, Hongfeng Zhan, Yuheng Jiao, Shihao Xu, Liang Yin, Jinsong Lan, Xiaoyong Zhu, Bo Zheng

\noindent\textbf{AI Infra and Engineering Contributors}: Zhixiang Zhao, Shuaiqi jia, Gaofeng Li, Yiquan Guo, Zhilong Hu, Dongzi Zhao, Yunfei Shen, Chong Ma

\noindent\textbf{Special Contributors}: (TaoSID2.0-MM-CF service Provider): Shuwen Xiao, Xiangheng Kong, and (TBStars-Item2Any Provider): Yuan Gao, Jun Song.

\section{Acknowledgement}
We would like to express our gratitude to the following teams for their invaluable contributions to this work:
\begin{itemize}
    \item The \textbf{TaoSID} team, for their support in the original product Semantic ID construction.
    \item The \textbf{TBStars} team, for providing product understanding knowledge (\textit{i.e.} item2any model).
    \item The \textbf{AI Infra} team, for inference acceleration and optimization during model deployment.
    \item The \textbf{Engineering Development} team and \textbf{Product} team, for their support in usage and product design throughout the online deployment process.
\end{itemize}

\clearpage
\bibliographystyle{plainnat}
\bibliography{refs}


\end{document}